\documentclass[conference]{IEEEtran}

\usepackage{graphicx}
\usepackage[draft=false,bookmarks=false]{hyperref} 
\hypersetup{
	colorlinks=true,       
	linkcolor=blue,        
	citecolor=blue,        
	filecolor=magenta,     
	urlcolor=blue         
}
\usepackage[utf8]{inputenc}  
\usepackage{amsmath}
\usepackage{subcaption}
\usepackage{xspace}
\usepackage{booktabs}
\usepackage{threeparttable}
\usepackage[inline]{enumitem}
\usepackage{mathtools}
\DeclarePairedDelimiter{\ceil}{\lceil}{\rceil}
\newcommand{\xgb}{\textsc{XGB}\xspace}
\newcommand{\axgb}{\textsc{AXGB}\xspace}
\newcommand{\axgba}{\textsc{AXGB$_A$}\xspace}
\newcommand{\axgbpush}{\textsc{AXGB$_{[p]}$}\xspace}
\newcommand{\axgbapush}{\textsc{AXGB$_{A[p]}$}\xspace}
\newcommand{\axgbrepl}{\textsc{AXGB$_{[r]}$}\xspace}
\newcommand{\axgbarepl}{\textsc{AXGB$_{A[r]}$}\xspace}
\newcommand{\bxgb}{\textsc{BXGB}\xspace}
\newcommand{\awedt}{\textsc{AWE-J48}\xspace}
\newcommand{\arf}{\textsc{ARF}\xspace}
\newcommand{\hatree}{\textsc{HAT}\xspace}
\newcommand{\lbht}{\textsc{LB$_{\text{HT}}$}\xspace}
\newcommand{\obht}{\textsc{OB$_{\text{HT}}$}\xspace}
\newcommand{\samknn}{\textsc{SAM}$_{\text{kNN}}$\xspace}
\newcommand{\rht}{\textsc{RHT}\xspace}
\newcommand{\adwin}{\textsc{ADWIN}\xspace}
\newcommand{\dsagra}{\textsf{AGR$_a$}\xspace}
\newcommand{\dsagrg}{\textsf{AGR$_g$}\xspace}
\newcommand{\dshyperf}{\textsf{HYPER$_f$}\xspace}
\newcommand{\dsseaa}{\textsf{SEA$_a$}\xspace}
\newcommand{\dsseag}{\textsf{SEA$_g$}\xspace}
\newcommand{\dsairl}{\textsf{AIRL}\xspace}
\newcommand{\dselec}{\textsf{ELEC}\xspace}
\newcommand{\dsweather}{\textsf{WEATHER}\xspace}
\newcommand{\refequation}[1]{Eq.~\ref{#1}}
\newcommand{\reffig}[1]{Figure~\ref{#1}}

\newcommand{\reftab}[1]{Table~\ref{#1}}
\newcommand{\refsec}[1]{Sec.~\ref{#1}}

\long\def\BEGINOMIT#1\ENDOMIT{\relax}  

\def\BibTeX{{\rm B\kern-.05em{\sc i\kern-.025em b}\kern-.08em
    T\kern-.1667em\lower.7ex\hbox{E}\kern-.125emX}}

\begin{document}

\title{Adaptive XGBoost for Evolving Data Streams
}
%
\author{
  \IEEEauthorblockN{Jacob Montiel\IEEEauthorrefmark{1},
                    Rory Mitchell\IEEEauthorrefmark{1},
                    Eibe Frank\IEEEauthorrefmark{1},
                    Bernhard Pfahringer\IEEEauthorrefmark{1},
                    Talel Abdessalem\IEEEauthorrefmark{2}
                    and Albert Bifet\IEEEauthorrefmark{1}\IEEEauthorrefmark{2}}
  \IEEEauthorblockA{\IEEEauthorrefmark{1}
                    Department of Computer Science, University of Waikato, Hamilton, New Zealand\\
                    Email: \{jmontiel, eibe, bernhard, abifet\}@waikato.ac.nz, r.a.mitchell.nz@gmail.com}
  \IEEEauthorblockA{\IEEEauthorrefmark{2}
                    LTCI, Télécom ParisTech, Institut Polytechnique de Paris, Paris, France\\
                    Email: talel.abdessalem@telecom-paris.fr}
}


\maketitle

\begin{abstract}
Boosting is an ensemble method that combines base models in a sequential manner to achieve high predictive accuracy. A popular learning algorithm based on this ensemble method is \textit{eXtreme Gradient Boosting} (\xgb). We present an adaptation of \xgb for classification of evolving data streams. In this setting, new data arrives over time and the relationship between the class and the features may change in the process, thus exhibiting concept drift. The proposed method creates new members of the ensemble from mini-batches of data as new data becomes available. The maximum ensemble size is fixed, but learning does not stop when this size is reached because the ensemble is updated on new data to ensure consistency with the current concept. We also explore the use of concept drift detection to trigger a mechanism to update the ensemble. We test our method on real and synthetic data with concept drift and compare it against batch-incremental and instance-incremental classification methods for data streams.
\end{abstract}


\section{Introduction}
\label{sec:intro}

The eXtreme Gradient Boosting (\xgb) algorithm is a popular method for supervised learning tasks. XGB is an ensemble learner based on boosting that is generally used with decision trees as weak learners. During training, new weak learners are added to the ensemble in order to minimize the objective function. Different to other boosting techniques, the complexity of the trees is also considered when adding weak learners: trees with lower complexity are preferred. 
Although configuring the multiple hyper-parameters in \xgb can be challenging, it performs at the state-of-the-art if this is done properly.

An emerging approach to machine learning comes in the form of learning from evolving data streams. It provides an attractive alternative to traditional batch learning in multiple scenarios. An example is fraud detection for online banking operations, where training is performed on massive amounts of data. In this case, consideration of runtime is critical: waiting for a long time until the model is trained means that potential frauds may pass undetected. Another example is the analysis of communication logs for security, where storing all logs is impractical (and in most cases unnecessary). The requirement to store all data is a significant limitation of methods that need to perform multiple passes over the data. 

Stream learning comprises a set of additional challenges, such as: models have access to the data only once and need to process it on the go since new data arrives continuously; models need to provide predictions at any moment in time; and there is a potential change in the relationship between features and learning targets, known as concept drift. Concept drift is a challenging problem, and is common in many real-world applications that aim to model dynamic systems. Without proper intervention, batch methods will fail after a concept drift because they are essentially trained for a different problem (concept). A common approach to deal with this phenomenon, usually signaled by the degradation of a batch model, is to replace the model with a new model, which implies a considerable investment on resources to collect and process data, train new models and validate them. In contrast, stream models are continuously updated and adapt to the new concept.



We list the contributions of our work as follows:
\begin{itemize}
    \item We propose an adaptation of the eXtreme Gradient Boosting algorithm for evolving data streams.
    \item We provide an open-source implementation\footnote{\href{https://github.com/jacobmontiel/AdaptiveXGBoostClassifier}{https://github.com/jacobmontiel/AdaptiveXGBoostClassifier}} of the proposed algorithm. 
    \item We perform a thorough evaluation of the proposed method in terms of performance, hyper-parameter relevance, memory, and training time.
    \item Our experimental results update the existing literature comparing instance-incremental and batch-incremental methods, with current state-of-the-art methods.
\end{itemize}

This paper is organized as follows: In Section~\ref{sec:related_work} we examine related work. The proposed method is introduced in Section~\ref{sec:axgb}. Section~\ref{sec:methodology} describes the methodology for our experiments. Results are discussed in Section~\ref{sec:results}. We present our conclusions in Section~\ref{sec:conclusions}. 

\section{Related Work}
\label{sec:related_work}


Ensemble methods are a popular approach to improve predictive performance and robustness. One of the first techniques to address concept drift with ensembles trained on streaming data is the \textit{SEA} algorithm \cite{street2001streaming}, a variant of bagging~\cite{Breiman1996Bagging} which maintains a fixed-size ensemble trained incrementally on chunks of data. \textit{Online Bagging}~\cite{Oza2005} is an adaptation of bagging for data streams. Similar to batch bagging, $M$ models are generated and then trained on $N$ samples. Different to batch bagging, where samples are selected with replacement, in Online Bagging, samples are assigned a weight based on $Poisson(1)$. \textit{Leveraging Bagging}~\cite{Bifet2010Leveraging} builds upon Online Bagging. The key idea is to increase accuracy and diversity on the ensemble via randomization. Additionally, Leveraging Bagging uses the \adwin~\cite{bifet2007ADWIN} drift detector; if a change is detected, the worst member in the ensemble is replaced by a new one.
The \textit{Ensemble of Restricted Hoeffding Trees}~\cite{Bifet2012RHT} combines the predictions of multiple tree models built from subsets of the full feature space using stacking. 
The \textit{Self Adjusting Memory} algorithm~\cite{Losing2017} builds an ensemble with models targeting current or former concepts. SAM works under the \textit{Long-Term} --- \textit{Short-Term} memory model (LTM-STM), where the STM focuses on the current concept and the LTM retains information about past concepts. 
\textit{Adaptive Random Forest}~\cite{Gomes2017ARF} is an adaptation of the Random Forest method designed to work on data streams. The base learners are Hoeffding Trees, attributes are randomly selected during training, and concept drift is tracked using \adwin on each member of the ensemble. 

In the batch setting, boosting is an extremely popular ensemble learning strategy. The \textit{Pasting Votes}~\cite{Breiman1999pasting} method is the first to apply boosting on large data sets by using different sub-sets of data for each boosting iteration; it does not require to store all data and potentially can be used on stream data. 
A similar approach is \textit{Racing Committees}~\cite{Frank2002racing}. Different to~\cite{Breiman1999pasting}, Racing Committees includes a adaptive pruning strategy to manage resources (time and memory). 
In the stream setting, a number of approaches for boosting have been proposed. \textit{Learn++.NSE}~\cite{Polikar2001learnpp}, inspired in \textit{AdaBoost}~\cite{Freund1997adaboost}, generates a batch-based ensemble of weak classifiers trained on different sample distributions and combines weak hypotheses through weighted majority voting. 

\BEGINOMIT
A boosting-like ensemble method explicitly designed to handle concept drift is proposed in~\cite{Scholz2007}. The ensemble is generated using knowledge-based sampling where patterns are discovered iteratively. 
\textit{Online Boost-By-Majority}~\cite{Beygelzimer2015obbm} is a stream version of the boost-by-majority algorithm~\cite{freund1995bbm} for binary classification, based on a weaker  assumption of online weak learners. 
The \textit{Online Multiclass Boost-By-Majority} algorithm~\cite{Jung2017OnlineMBBM} is an extension for multiclass classification. 
\textit{BoostVHT}~\cite{Vasiloudis2017boostvht} improves training time by taking advantage of parallel and distributed architectures without modifying the order assumption in boosting. To achieve this, they use as base learner the Vertical Hoeffding Trees\cite{kourtellis2016vht}, a model-parallel distributed version of the Hoeffding Tree.
\ENDOMIT

In stream learning, two main branches of algorithms can be distinguished depending on the schema used to train a model. \textit{Instance-incremental} methods~\cite{Oza2005,Bifet2010Leveraging,Bifet2012RHT,Gomes2017ARF,Losing2017}, where a single sample is used at a time, and \textit{batch-incremental} methods~\cite{Breiman1999pasting,Polikar2001learnpp,Frank2002racing} that use batches of data: Once a given number of samples are stored in the batch, they are used to train the model. The \textit{Accuracy-Weighted Ensembles}~\cite{Wang2003AWE}, is a framework for mining streams with concept drift using weighted classifiers. Members of the ensemble are discarded by an instance-based pruning strategy if they are below a confidence threshold. A relevant study is~\cite{Read2012}, where the authors compare batch-incremental and instance-incremental methods for the task of classification of evolving data streams. While instance-incremental methods perform better on average, batch-incremental methods achieve similar performance in some scenarios.

\section{Adapting XGB for Stream Learning}
\label{sec:axgb}
In this section, we present an adaptation of the XGB algorithm \cite{Chen2016XGB} suitable for evolving data streams.

\subsection{Preliminaries}
The goal of supervised learning is to predict the responses $Y=\{y_i\}: i \in \{1,2,\ldots,n\}$ corresponding to a set of feature vectors $X=\{\vec{x}_i\}: i \in \{1,2,\ldots,n\}$. Ensemble methods yield predictions $\hat{y_i}$ corresponding to a given input $\vec{x}_i$ by combining the predictions of all the members of the ensemble $\mathrm{E}$. In this paper, we focus on binary classification, that is, $y \in \{C_1,C_2\}$.

In the case of boosting, the ensemble $\mathrm{E}$ is created sequentially. In each iteration $k$, a new base function $f_k$ is selected and added to the ensemble so that the loss $\ell$ of the ensemble is minimized:
\begin{equation}\label{eq:objective}
    \ell(\mathrm{E}) = \sum_{k=1}^{K}\ell(Y,\hat{Y}^{(k-1)} + f_k(X)) + \Omega(f_k).
\end{equation}
Here, $K$ is the number of ensemble members and each $f_k \in \mathcal{F}$ with $\mathcal{F}$ being the space of possible base functions. Commonly, this is the space of regression trees, so each base function is a step-wise constant predictor and the ensemble prediction $\hat Y$, which is simply the sum of all $K$ base functions, is also step-wise constant. The regularization parameter $\Omega$ penalizes complex functions.

The ensemble is created using forward additive modeling, where new trees are added one at a time. At step $k$, the training data is evaluated on existing members of the ensemble and the corresponding prediction scores $Y^{(k)}$ are used to drive the creation of new members of the ensemble. The base functions predictions are combined additively:

\begin{align}\label{eq:ensemble_creation}
\begin{split}
    \hat{Y}^{(k)} &= \sum_{k=1}^K f_k(X)= \hat{Y}^{(k-1)} + f_k(X)
\end{split}
\end{align}
The final prediction for a sample $\hat{y}_i$ is the sum of the predictions for each tree $f_{k}$ in the ensemble.
\begin{equation}
    \hat{y_i} = \sum_{k=1}^{K}f_{t}(x_i)
\end{equation}

\subsection{Adaptive eXtreme Gradient Boosting}\label{sec::proposed_method}
In the batch setting, \xgb training is performed using all available data $(X, Y)$. However, in the stream setting, we receive new data over time, and this data may be subject to concept drift. A continuous stream of data can be defined as $A=\{(\vec{x}_t,y_t)\} | t = 1,\ldots,T$ where $T \rightarrow \infty$, $\vec{x}_t$ is a feature vector, and $y_t$ the corresponding target. We now describe a modified version of the \xgb algorithm for this scenario, called \textsc{\underline{A}daptive e\underline{X}treme \underline{G}radient \underline{B}oosting} (\axgb).  \axgb uses an incremental strategy to create the ensemble. Instead of using the same data to select each base function $f_i$, it uses sub-samples of data obtained from non-overlapping (tumbling) windows. More specifically, as new data samples arrive, they are stored in a buffer $w=(\vec{x}_i, y_i): i \in \{1,2,\dots,W\}$ with size $|w|=W$ samples. Once the buffer is full, \axgb proceeds to train a single $f_{k}$. We can rewrite \refequation{eq:ensemble_creation} as:

\begin{align}\label{eq:ensemble_creation_axgb}
\begin{split}
    \hat{Y}^{(k)} &= \sum_{k=1}^K f_k(w_k)= \hat{Y}^{(k-1)} + f_k(w_k)
\end{split}
\end{align}

The index $k$ of the new base function within the ensemble determines the way in which this function is obtained. If it is the first member of the ensemble, $f_1$, then the data in the buffer is used directly. If $k>1$, then the data is passed through the ensemble and the residuals from the first $k-1$ models in the ensemble are used to obtain the new base function $f_k$.

\subsection{Updating the Ensemble}\label{sec::ensemble_creation}
\begin{figure}[!t]
    \begin{subfigure}[b]{0.5\textwidth}
    \centering
        \includegraphics[height=3cm]{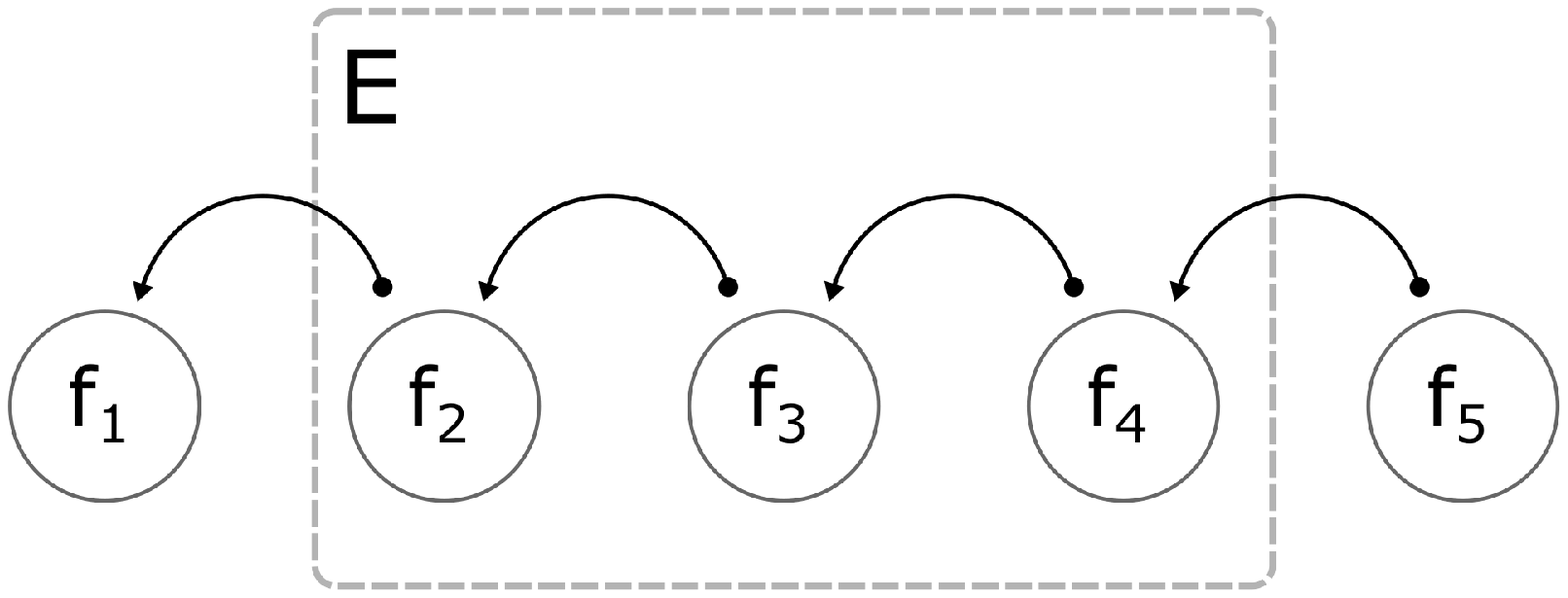}
        \caption{Push}
        \label{fig:push}
    \end{subfigure}
    \\
    \begin{subfigure}[b]{0.5\textwidth}
        \centering
        \includegraphics[height=3cm]{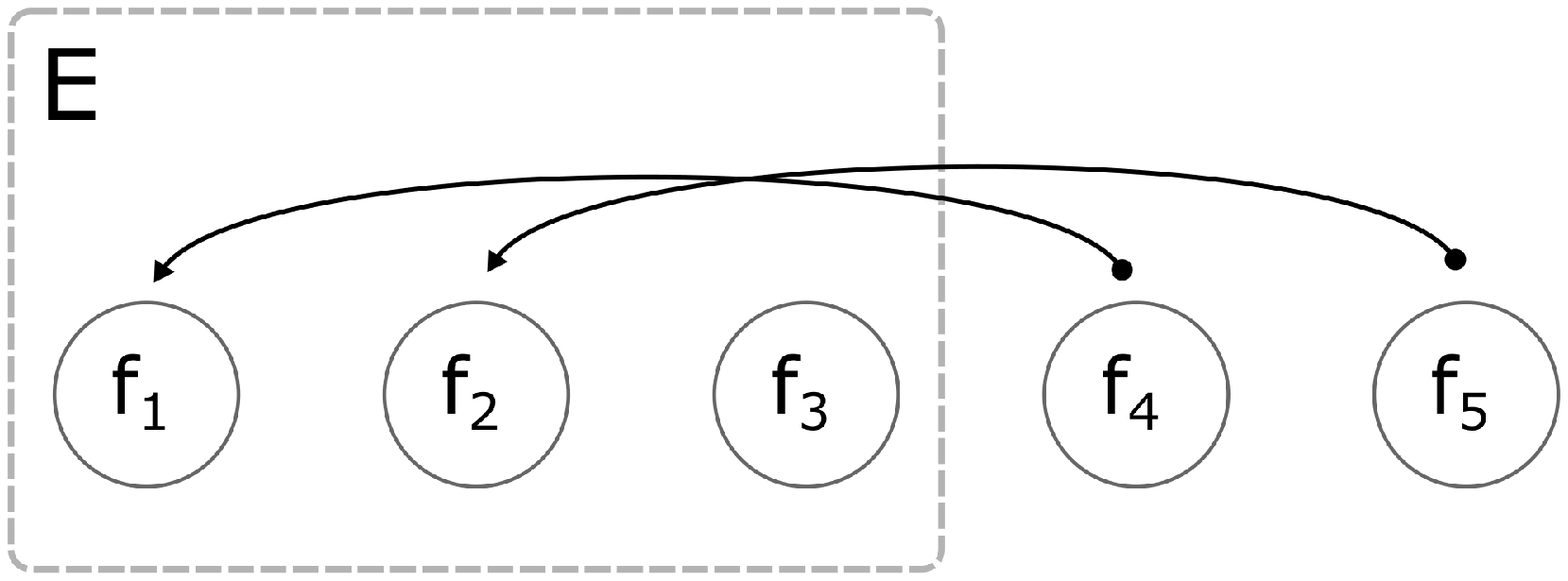}
        \caption{Replace}
        \label{fig:replace}
    \end{subfigure}
    \caption{Ensemble creation strategies.}\label{fig:strategies}
\end{figure}

Given that data streams are potentially infinite and may change over time, learned predictors must be updated continuously. Thus, it is essential to define a strategy to keep the \axgb ensemble updated once it is full. In the following, we consider two strategies for this purpose:
\begin{itemize}
    \item A \textit{push} strategy (\axgbpush), shown in \reffig{fig:push}, where the ensemble resembles a queue. When new models are created they are appended to the ensemble. If the ensemble is full then older models are removed before appending a new model.
    \item A \textit{replacement} strategy (\axgbrepl), shown in \reffig{fig:replace}, where older members of the ensemble are replaced with newer ones.
\end{itemize}

Notice that in both cases, we have to wait $K$ iterations to have a completely new ensemble. However, in \axgbrepl, newer models have a more significant impact on predictions than older ones, while the reverse is true for \axgbpush.

A requirement in stream learning is that models are ready to provide predictions at any time. Given the incremental nature of \axgb, if the window (buffer) size $W$ is fixed, the ensemble will require $K \cdot W$ samples to create the full ensemble. A negative aspect of this approach is that performance can be sub-optimal at the beginning of the stream. To overcome this, \axgb uses a dynamic window size $W$ that doubles in each iteration from a given minimum size $W_{min}$ until a maximum size $W_{max}$ is reached. In other words, it grows exponentially until reaching $W_{max}$. The window size, $W(i)$, for the $i$th iteration is defined as:

\begin{equation}\label{eq:window_size}
    W(i)=min(W_{min}\cdot 2^{i}, W_{max})
\end{equation}

From \refequation{eq:window_size}, we see that the number of iterations $i$ required to reach the maximum window size is:

\begin{equation}
    i = \ceil[\bigg]{\log_2\left(\frac{W_{max}}{W_{min}}\right)}
\end{equation}

Similarly, we see that the number of samples required to create $K$ models to fill the ensemble is smaller when using the dynamic window size approach than when using a fixed window size $W_{max}$ given that
 \begin{equation}
     \sum_{i=0}^{K-1}W_{min}\cdot2^{i} \ll K\cdot W_{max}.
 \end{equation}

Because we monotonically increase the window size, we see that both ensemble update strategies replace base functions trained on small windows with newer ones trained on more data. 

\subsection{Handling Concept Drift}
Although the incremental strategy used by \axgb to create the ensemble indirectly deals with concept drift---new members of the ensemble are added based on newer data---it may be too slow to adjust to fast drifts. Hence, we use \adwin \cite{bifet2007ADWIN}, to track changes in the performance of \axgb, as measured by a metric such as classification accuracy. We use subscript $_A$ to denote ADWIN, therefore the concept-drift-aware version of \axgb is referred in the following as \axgba.

\axgba uses the change detection signal obtained from \adwin to trigger a mechanism to update the ensemble. This mechanism works as follows:

\begin{enumerate}
    \item Reset the size of the window $w$ to the defined minimum size $W_{min}$. 
    \item Train and add new members to the ensemble depending on the chosen strategy:
    \begin{enumerate}
        \item Push: New ensemble members are appended to the ensemble while the oldest are removed from it. Since new models are trained on increasing window sizes they will be added at a faster rate initially; this effectively works as a \textit{flushing} strategy to update the ensemble.
        \item Replacement: The index used to replace old members of the ensemble is reset so that it points to the beginning of the ensemble. There are two considerations: First, new models replace the oldest ones in the ensemble. Second, new models are trained without considering the residuals of old models that were trained on the older concept.
    \end{enumerate}
\end{enumerate}


\section{Experimental Methodology}
\label{sec:methodology}
In this section, we describe the methodology of our tests, which we classify into the following categories: predictive performance, hyper-parameter relevance, memory usage / model complexity,
and training time.

\begin{enumerate}
    \item \textbf{Predictive performance}. Our first set of tests evaluates the predictive performance of \axgb. For this we use both, synthetic and real-world data sets. We then proceed to compare \axgb against other learning methods. This comparison is defined by the nature of the learning method as follows:
    \begin{enumerate}
        \item \textit{Batch-incremental methods}. In this type of learning methods, batches of samples are used to incrementally update the model. We compare \axgb against a batch-incremental ensemble created by combining multiple per-batch base models. New base models are trained independently on disjoint batches of data (windows). When the ensemble is full, older models are replaced with newer ones. Predictions are formed by majority vote. In order to compare this approach with \axgb, we use \xgb as the base batch-learner to learn an ensemble for each batch. Thus, our batch-incremental model is an ensemble of \xgb ensembles. We refer to this batch-incremental method as \bxgb. We also consider Accuracy-Weighted Ensembles with Decision Trees as the base batch-learner. We refer to this method as \awedt. We choose this configuration since \awedt is reported as the top batch-incremental performer in \cite{Read2012}, so it serves as a baseline for batch-incremental methods.
        \item \textit{Instance-incremental methods}. We are also interested in comparing \axgb against methods that update their model one instance at a time. The following instance-incremental methods are used in our tests: Adaptive Random Forest~(\arf), Hoeffding Adaptive Tree~(\hatree), Leverage Bagging with Hoeffding Tree as base learner~(\lbht), Oza Bagging with Hoeffding Tree as base learner~(\obht), Self Adjusting Memory with kNN (\samknn) and the Ensemble of Restricted Hoeffding Trees~(\rht). In \cite{Read2012}, \lbht is reported as the top instance-incremental performer.
    \end{enumerate}
    We perform non-parametric tests to verify whether there are statistically significant differences between algorithms, as described in~\cite{Demsar2006StatisticalComparison, Garcia2009ExtStatisticalComparison}.
    \item \textbf{Hyper-parameter relevance}. The \xgb algorithm relies on multiple hyper-parameters, which can make the model hard to tune for different problems. We are interested in analyzing the impact of hyper-parameters in \axgb. For this purpose, we use a hyper parameter tuning setup where a model is trained on the first 30\% of the data stream using different combinations of hyper-parameters. Then, the best performers during the training phase are evaluated on the remaining 70\% of the stream. To evaluate the influence of hyper-parameters, we compare performance between \axgb and \bxgb.
    \item \textbf{Memory usage and model complexity}. The potentially infinite number of samples in data streams requires resources such as time and memory to be properly managed. We use the total number of nodes in the ensemble to gain insight into memory usage and  model complexity as \axgb is trained on a data stream. We compare the proposed versions of \axgb against a baseline \xgb model trained on all the data from the stream. The baseline number of nodes in the \xgb model is expected to be larger than the number of nodes in incremental-models that evolve with the stream. By analyzing memory usage and model complexity we aim to get intuition on the evolution of the model over time.
    \item \textbf{Training time}. Another relevant way to analyze the proposed method is in terms of training time. We compare the training time of the different versions of \axgb against \xgb, reporting results in terms of training time (seconds) and in terms of throughput (samples per second).
\end{enumerate}

Our implementation of \axgb is based on the official \xgb C-API\footnote{\url{https://github.com/dmlc/xgboost}} on top of \textsf{scikit-multiflow}\footnote{\url{https://github.com/scikit-multiflow/scikit-multiflow}} \cite{skmultiflow}, a stream learning framework in Python. Tests are performed using the official \xgb implementation, the implementations of \arf, \rht and \textsc{AWE} in \textsf{MOA} \cite{MOA}, and for the rest of the methods, the implementations available in \textsf{scikit-multiflow}. Default parameters of the algorithms are used unless otherwise specified.

\subsection{Data}\label{sec:data}

In the following, we provide a short description of the synthetic and real world datasets used in our tests. All datasets are publicly available. A summary of the datasets used in our experiments is available in \reftab{tab:datasets}.

\begin{itemize}
    \item \textit{AGRAWAL} -- Based on the Agrawal generator~\cite{aggarwal2003framework}, represents a data stream with six nominal and three numerical features. Different functions map instances into two different classes. Three abrupt drifts are simulated for \dsagra and three gradual drifts for \dsagrg.
    \item \textit{HYPER} -- A stream with fast incremental drifts where a $d$-dimensional hyperplane changes position and orientation. Obtained from a random hyperplane generator~\cite{hulten2001mining}.
    \item \textit{SEA} -- A data stream with three numerical features where only two attributes are related to the target class. Created using the SEA generator~\cite{street2001streaming}. Three abrupt drifts are simulated for \dsseaa~and three gradual drifts for \dsseag.
    \item \textit{AIRLINES} -- Real world data containing information from scheduled departures of commercial flights within the US. The objective is to predict if a flight will be delayed.
    \item \textit{ELECTRICITY} -- Data from the Australian New South Wales Electricity Market, where prices are not fixed but change based on supply and demand. The two target classes represent changes in the price (up or down).
    \item \textit{WEATHER} -- Contains weather information collected between 1949--1999 in Bellevue, Nebraska. The goal is to predict rain on a given date.
\end{itemize}

\begin{table}[t]
    \centering
    \scriptsize
    \caption{Datasets. [Type] S:~synthetic data; R:~real world data. [Drifts] A:~abrupt, G:~gradual; I$_f$:~incremental fast, ?:~drifts with unknown nature.}\label{tab:datasets}
    \begin{tabular}{@{}lrcccc@{}}
        \toprule
        Dataset              & \multicolumn{1}{c}{\# instances} & \# features & \# classes & Type & Drift \\ \midrule
        \textsf{AGR$_a$}     & 1000000                          & 9           & 2          & S    & A     \\
        \textsf{AGR$_g$}     & 1000000                          & 9           & 2          & S    & G     \\
        \textsf{HYPER$_f$}   & 1000000                          & 10          & 2          & S    & I$_f$ \\
        \textsf{SEA$_a$}     & 1000000                          & 3           & 2          & S    & A     \\
        \textsf{SEA$_g$}     & 1000000                          & 3           & 2          & S    & G     \\
        \textsf{AIRL}        & 539383                           & 7           & 2          & R    & ?     \\
        \textsf{ELEC}        & 45312                            & 6           & 2          & R    & ?     \\
        \textsf{WEATHER}     & 18159                            & 8           & 2          & R    & ?     \\
        \bottomrule
    \end{tabular}
\end{table}


\section{Experimental Results}
\label{sec:results}
The results discussed in this section provide information about predictive performance, parameter relevance, memory and time for the different versions of \axgb. 

\subsection{Predictive Performance}\label{sec:results_performance}
We evaluate the performance of \axgb against other batch-incremental methods and against instance-incremental methods. We use prequential evaluation \cite{dawid1984prequential}, where predictions are generated for a sample in the stream before using it to train/update the model. We use classification accuracy as the metric in our tests in order to measure performance. First, we compare the different versions of \axgb (\axgbpush, \axgbapush, \axgbrepl and \axgbarepl) against two batch-incremental methods: \bxgb and \awedt. The parameters used to configure these methods are available in~\reftab{tab:parameters}.

\begin{table}[!t]
    \centering
    \caption{Parameters used for batch-incremental methods. }\label{tab:parameters}
    \begin{threeparttable}
        \begin{tabular}{@{}lrrr@{}}
            \toprule
            Parameter                        & \axgb\tnote{*} & \bxgb & \awedt \\ \midrule
            ensemble size                    & 30    & 30    & 30     \\
            ensemble size (base learner)     & -     & 30    & -      \\
            max window size                  & 1000  & 1000  & 1000   \\
            min window size                  & 1     & -     & -      \\
            max depth                        & 6     & 6     & -      \\
            learning rate                    & 0.3   & 0.3   & -      \\ \bottomrule
        \end{tabular}
        \begin{tablenotes}
            \item[*] \scriptsize{The same parameter configuration is used for all variations: \axgbpush, \axgbapush, \axgbrepl and \axgbarepl.}
        \end{tablenotes}
    \end{threeparttable}
\end{table}

Results comparing against batch-incremental methods are available in \reftab{tab:testbatchinc}. We see that the overall top performer in this test is \axgbrepl, followed by \axgbarepl. Next are the versions of \axgb using the push strategy. Interestingly, we find that \awedt performs better than \bxgb, which comes last in this test. This is noteworthy considering that the base learner in \awedt (a single decision tree) is simpler than the one in \bxgb (an ensemble of trees generated using XGBoost).
\begin{table}[!b]
    \centering
    \caption{Comparing performance of \axgb vs batch-incremental methods.}\label{tab:testbatchinc}
    \resizebox{.5\textwidth}{!}{
    \begin{tabular}{lcccccc}
        \toprule
        Dataset           & \axgbpush  & \axgbrepl       & \axgbapush  & \axgbarepl      & \bxgb           & \awedt  \\ \midrule
        \dsagra           & 0.919      & \textbf{0.931}  & 0.927       & 0.928           & 0.703           & 0.926   \\
        \dsagrg           & 0.896      & \textbf{0.907}  & 0.897       & 0.901           & 0.710           & 0.905   \\
        \dsairl           & 0.604      & 0.621           & 0.611       & 0.618           & \textbf{0.641}  & 0.599   \\
        \dselec           & 0.718      & 0.739           & 0.740       & \textbf{0.747}  & 0.702           & 0.614   \\
        \dshyperf         & 0.822      & \textbf{0.847}  & 0.825       & \textbf{0.847}  & 0.756           & 0.777   \\
        \dsseaa           & 0.865      & \textbf{0.875}  & 0.866       & 0.874           & 0.856           & 0.860   \\
        \dsseag           & 0.863      & \textbf{0.873}  & 0.863       & 0.872           & 0.857           & 0.860   \\
        \dsweather        & 0.765      & \textbf{0.774}  & 0.767       & 0.747           & 0.737           & 0.712   \\ [5pt]
        \textbf{avg. rank}& 4.188      & \textbf{1.438}  & 3.063       & 2.313           & 5.125           & 4.875   \\ \bottomrule
    \end{tabular}
    }
\end{table}

These tests provide insights into the different versions of \axgb. We see that, in the push-strategy versions, tracking performance to detect concept drift (\axgbapush) provides a consistent advantage over the drift-unaware approach (\axgbpush). The reason for this is that, as expected, \axgbapush reacts faster to changes in performance: When a drift is detected, the window size is reset and new models are quickly added to the ensemble, flushing-out older models.
This is not the case for methods using the replace-strategy, with \axgbrepl providing the best performance for most datasets. These results are significant given the compromise between the computational overhead of tracking concept drift and the gains in performance. We analyze this trade-off when discussing results of the running time tests.

Next, we compare \axgb against instance-incremental methods. Results are shown in \reftab{tab:testinstinc}. For \axgb, we only show results of \axgbrepl and \axgbarepl. We see that the top performer in this test is \arf, closely followed by \rht. \axgb's performance is not on par with that of the top performers, but it is important to note that 
\begin{enumerate*}[label=(\roman*)]
\item these results are consistent with those in \cite{Read2012}, where instance-incremental methods outperform batch-incremental methods, and 
\item both \axgbrepl and \axgbarepl are placed in the top tier between \lbht and \samknn.
\end{enumerate*}

\begin{figure}[!t]
    \centering
    \includegraphics[width=.49\textwidth]{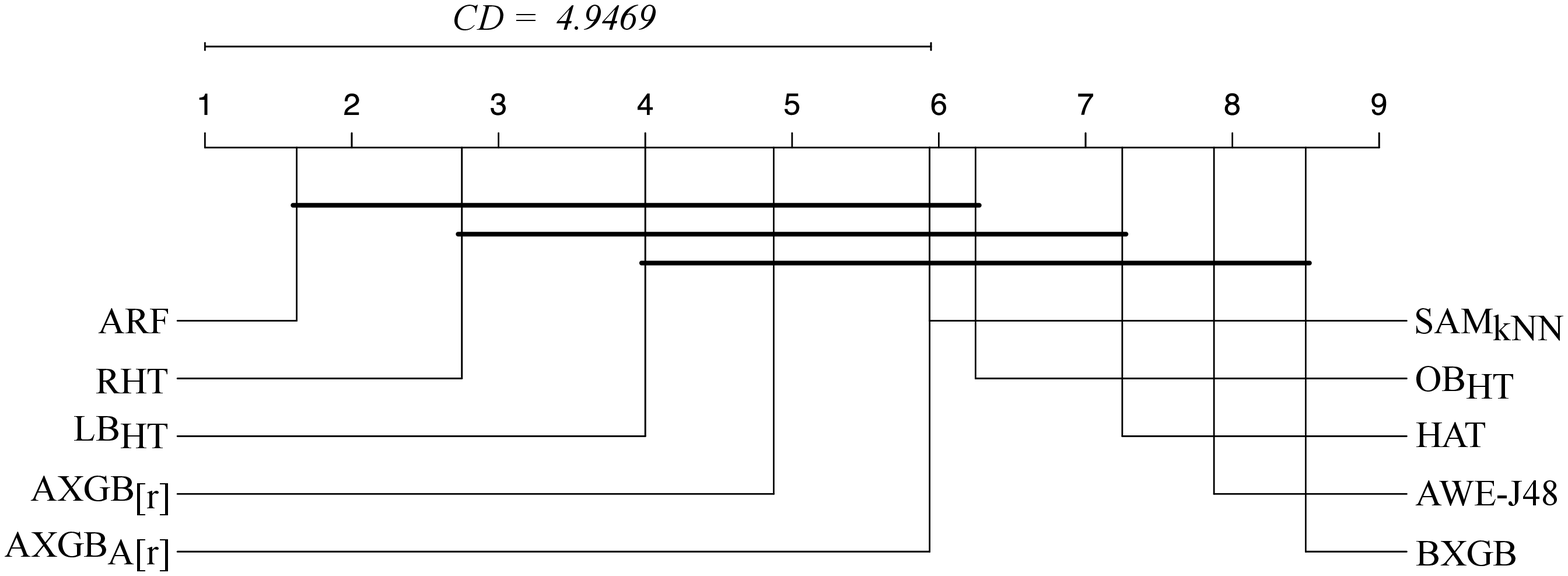}
    \caption{Nemenyi post-hoc test (95\% confidence level), identifying statistical differences between all methods in our tests. }\label{fig:cd_plot}
\end{figure}

The corrected Friedman test with $\alpha=0.05$ indicates that there are statistical significant differences between the methods in~\reftab{tab:testbatchinc} and~\reftab{tab:testinstinc} . The follow-up post-hoc Nemenyi test,~\reffig{fig:cd_plot}, indicates that there are no significant differences among the methods in the top tier. We believe that these findings serve to indicate the potential of eXtreme Gradient Boosting for data streams.

\begin{table}[t]
    \centering
    \caption{Comparing performance of \axgb vs instance-incremental methods.}\label{tab:testinstinc}
    \resizebox{.5\textwidth}{!}{
    \begin{tabular}{lcccccccc}
        \toprule
        Dataset            & \axgbrepl  & \axgbarepl  & \arf            & \hatree         & \lbht  & \obht  & \samknn  & \rht           \\ \midrule
        \dsagra            & 0.931      & 0.928       & \textbf{0.939}  & 0.807           & 0.881  & 0.915  & 0.686    & 0.936          \\
        \dsagrg            & 0.907      & 0.901       & \textbf{0.912}  & 0.792           & 0.858  & 0.847  & 0.669    & 0.911          \\
        \dsairl            & 0.621      & 0.618       & \textbf{0.680}  & 0.608           & 0.670  & 0.658  & 0.605    & 0.648          \\
        \dselec            & 0.739      & 0.747       & 0.855           & \textbf{0.874}  & 0.836  & 0.794  & 0.799    & 0.873          \\
        \dshyperf          & 0.847      & 0.847       & 0.849           & 0.869           & 0.814  & 0.806  & 0.870    & \textbf{0.896} \\
        \dsseaa            & 0.875      & 0.874       & \textbf{0.897}  & 0.827           & 0.891  & 0.869  & 0.876    & 0.889          \\
        \dsseag            & 0.873      & 0.872       & \textbf{0.893}  & 0.825           & 0.889  & 0.869  & 0.873    & 0.885          \\
        \dsweather         & 0.774      & 0.747       & \textbf{0.791}  & 0.693           & 0.783  & 0.749  & 0.781    & 0.758          \\ [5pt]
        \textbf{avg. rank} & 4.750      & 5.688       & \textbf{1.625}  & 6.125           & 3.750  & 6.000  & 5.313    & 2.750          \\ \bottomrule
    \end{tabular}
    }
\end{table}

\subsection{Hyper-parameter Relevance}\label{sec:results_tuning}

As previously mentioned, hyper parameters play a key role in the performance of \xgb. Thus, we also need to consider their impact in \axgb. In order to do so, we present results obtained by running multiple tests on different combinations of key parameters: the maximum depth of the trees, the learning rate (eta), the ensemble size, and the maximum and minimum window size. To cover a wide range of values for each parameter, we use a grid search based on the grid parameters specified in \reftab{tab:parameter-grid}. The parameter grid corresponds to a total of $4\times4\times5\times5\times3=1200$ combinations. For this test, we compare the following \xgb-based methods: \axgbpush, \axgbapush and \bxgb. 

\begin{table}[!t]
    \centering
    \caption{Parameter grid used to evaluate hyper-parameters relevance.}\label{tab:parameter-grid}
    \centering
    \begin{tabular}{@{}lc@{}}
        \toprule
        Parameter       & Values                      \\ \midrule
        max depth       & 1, 5, 10, 15                \\
        learning rate   & 0.01, 0.05, 0.1, 0.5        \\
        ensemble size   & 5, 10, 25, 50, 100          \\
        max window size & 512, 1024, 2048, 4096, 8192 \\
        min window size & 4, 8, 16                    \\ \bottomrule
    \end{tabular}
\end{table}

For establishing the effect of parameter tuning, the test is split into two phases: training and optimization on the first 30\% of the stream---using this validation data to evaluate all parameter combinations in the grid and choosing the best one using prequential evaluation of classification accuracy---and performance evaluation on the remaining 70\% of the stream to establish accuracy of the parameter-optimized algorithm by evaluating the algorithm with the identified parameter settings using prequential evaluation on this remaining data. The ensemble model is trained from scratch in this second phase. This strategy is limited in the sense that the nature of the validation data, including concepts drifts, is assumed to be similar to that of the remaining data. Nonetheless, it provides insights into the importance of hyper parameters. 

Results from this experiment are available in \reftab{tab:testtuning}. Reported results correspond to measurements obtained with parameter tuning (\textit{Tuning}) vs. reference results (\textit{Ref}) obtained using the fixed parameters in \reftab{tab:parameters}, building an ensemble from scratch on the same 70\% portion of the stream. 

\begin{table}[!b]
    \centering
    \caption{Parameter tuning results.}\label{tab:testtuning}
    \resizebox{.5\textwidth}{!}{
    \begin{tabular}{lcccccc}
        \toprule
                    & \textit{Ref}  & \textit{Tuning} & \textit{Ref}    & \textit{Tuning}  & \textit{Ref}   & \textit{Tuning} \\
        Dataset     & \axgbpush     & \axgbpush       & \axgbapush      & \axgbapush       & \bxgb          & \bxgb           \\ \midrule
        \dsagra     & 0.881         & 0.927           & \textbf{0.933}  & 0.931            & 0.727          & 0.930           \\
        \dsagrg     & 0.898         & 0.906           & 0.902           & 0.905            & 0.728          & \textbf{0.909}  \\
        \dsairl     & 0.616         & 0.627           & 0.588           & 0.628            & 0.632          & \textbf{0.639}  \\
        \dselec     & 0.713         & 0.736           & 0.658           & 0.739            & 0.631          & \textbf{0.742}  \\
        \dshyperf   & 0.816         & 0.873           & 0.833           & 0.876            & 0.754          & \textbf{0.904}  \\
        \dsseaa     & 0.879         & 0.889           & 0.881           & \textbf{0.892}   & 0.854          & 0.890           \\
        \dsseag     & 0.877         & 0.888           & 0.878           & \textbf{0.889}   & 0.855          & 0.888           \\
        \dsweather  & 0.755         & 0.767           & 0.758           & 0.765            & 0.703          & \textbf{0.782}  \\ [5pt]
        average     & 0.804         & 0.827           & 0.804           & 0.828            & 0.736          & \textbf{0.835}  \\ \bottomrule
    \end{tabular}
    }
\end{table}

We can see that optimizing hyper parameters clearly benefits all methods. As expected, hyper-parameters can provide an advantage over other methods. In this case, under-performers are now on par or above \lbht. Surprisingly, \bxgb obtains the largest boost in performance and is now the method that performs best. When analyzing the parameter configurations (detailed results not included due to space constraints), we see that \bxgb favors smaller values for max window size, learning rate and max depth. The observed increase in performance can be attributed to the impact of the hyper parameters on the base learner in \bxgb (batch \xgb models), remembering that \bxgb is an ensemble of ensembles. Another factor to consider is the small window sizes. In practice, having smaller windows means that models are replaced faster as the stream progresses and this can ameliorate the lack of drift awareness to some degree. It is reasonable that the same applies to the reduction in the performance gap between \axgbpush and \axgbapush. In the case of \axgb, our results show that the learning rate has a consistent impact on performance (lower is better), followed by max window size and max depth. Finally, our tests reveal the contrast in the impact of the ensemble size on the two versions of \axgb. While \axgbpush benefits from a smaller ensemble size, the opposite applies to \axgbapush. This supports the intuition that drift-aware methods can benefit from larger ensembles (to build complex models) which adapt faster in the presence of drift by triggering the corresponding ensemble update mechanism. On the other hand, batch-incremental methods without explicit drift detection mechanisms rely on their natural ability to adapt, which can be counterproductive with large ensemble sizes. It is important to note that although \bxgb is the top performer in this test, it is not efficient in terms of resources (time and memory), which affects stream 
applications where resources are limited.



\subsection{Memory Usage and Model Complexity}\label{sec:results_memory}
\begin{figure}[!t]
    \centering
    \begin{subfigure}[b]{0.49\textwidth}
        \centering
        \includegraphics[width=\textwidth]{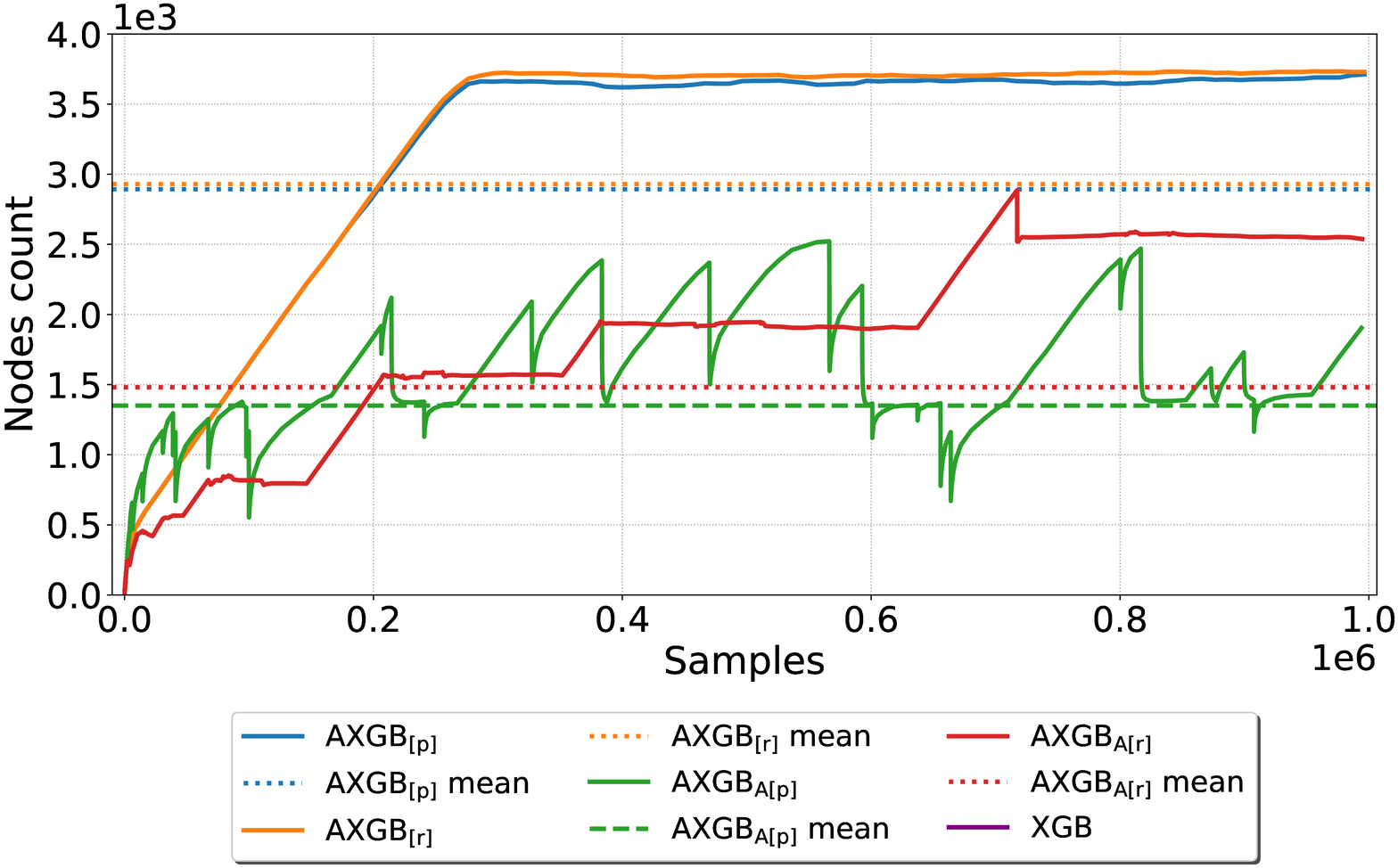}
        \vspace{-12mm}
        \caption{Total nodes in the ensemble.}\label{fig:nodes}
    \end{subfigure}
    \\
    \begin{subfigure}[b]{0.49\textwidth}
        \centering
        \includegraphics[width=\textwidth]{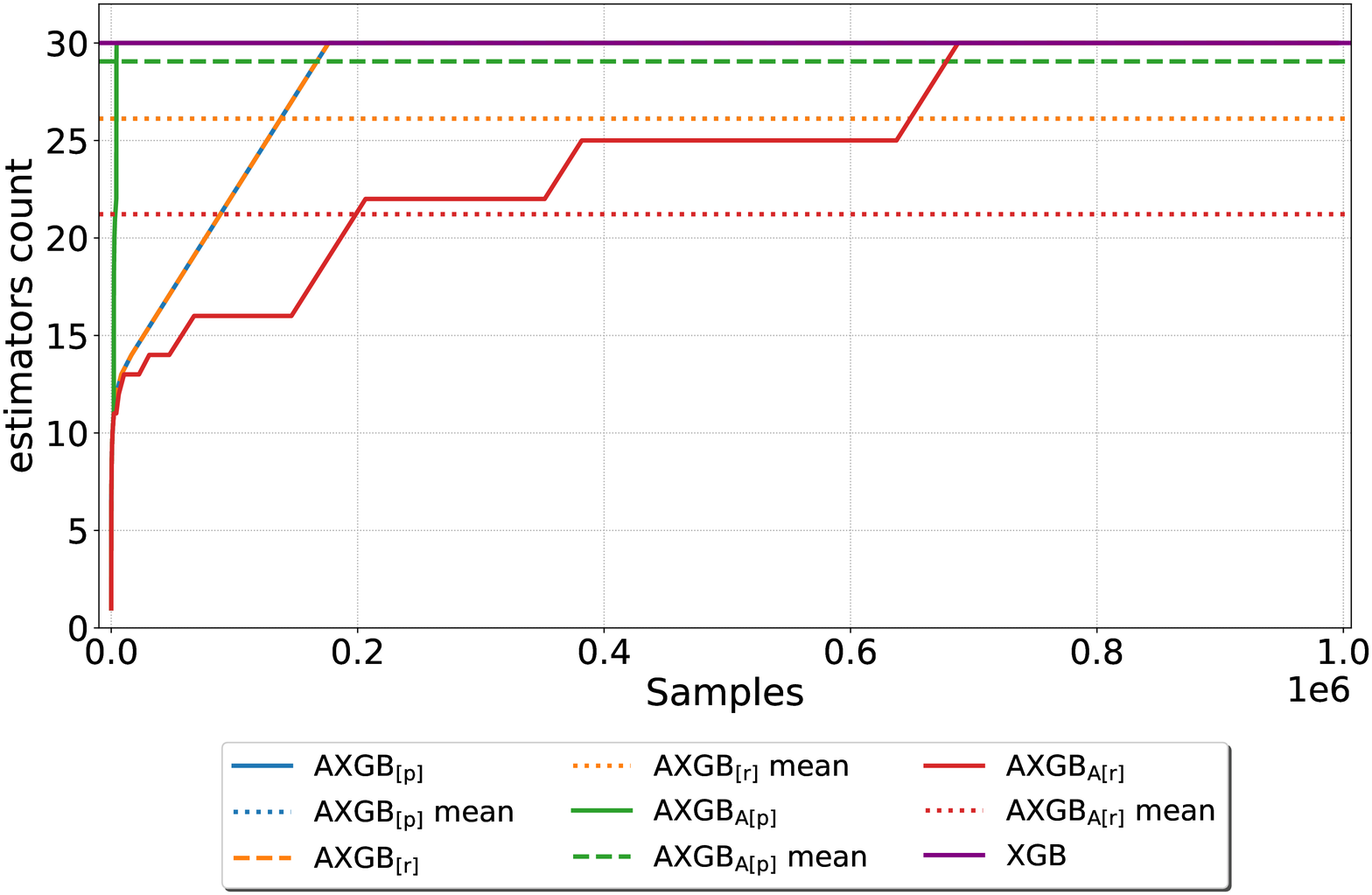}
        \vspace{-10mm}
        \caption{Number of ensemble members.}\label{fig:ensemble_size}
    \end{subfigure}
    \caption{Insight into ensemble complexity by number of nodes and ensemble members over the stream. For reference, an \xgb model with 30 ensemble members trained on all the data has 12.7K nodes (outside the plot area).}\label{fig:model_complexity}
\end{figure}

In this section, we analyze memory usage of the proposed methods during the learning process. For this purpose, we count the number of nodes in the ensemble, including both leaf nodes and internal nodes of each tree. This approach also provides some intuition regarding the model's complexity. We perform this test on a synthetic dataset with 40 features (only 30 of which are informative) and 5\% noise, corresponding to the \textit{Madelon} dataset, described in \cite{Guyon2007}.
We use 1 million samples for training and calculate the number of nodes in the ensemble to get an estimate of the model size. Models are trained using the following configuration: ensemble size~=~30, max window size~=~10K, min window size~=~1, learning rate~=~0.05, max depth~=~6. We measure the number of nodes as new members of the ensemble are introduced. Results from this test are available in \reffig{fig:model_complexity}, and serve to compare the number of nodes in the batch model vs. the stream models. For reference, the number of nodes in the \xgb model is 12.7K (outside the plot area). It is important to note that this number is constant since it represents the size of the model trained on all the data. 

In \reffig{fig:nodes}, we see that \axgbpush and \axgbrepl have similar behaviors. As the stream progresses, the number of nodes added to the ensemble increases until reaching a plateau. This is expected since new models are trained on larger windows of data. The plateau corresponds to the region where the ensemble is complete and old members of the ensemble are replaced by new members trained on equally large windows. On the other hand, \axgbapush and \axgbarepl also exhibit an incremental increase in the number nodes over the stream---at a lower rate than the \axgb versions---but with some interesting differences. In \axgbapush, we see multiple drop points in the nodes count, which can be attributed to the ensemble update mechanism. When drift is detected, the window size is reset and new models are pushed into the ensemble, in other words, simpler models are quickly introduced into the ensemble. In contrast, the number of nodes in \axgbarepl increases steadily. This difference in number of nodes can explain the difference in performance between \axgbrepl and \axgbarepl discussed in \refsec{sec:results_performance}.

We also analyze \axgb by counting the number of models in the ensemble across the stream, shown in \reffig{fig:ensemble_size}. Notice that the number of models reach the maximum value when the ensemble is full; from that point on, new models replace old ones. As anticipated, we see that \axgbapush fills the ensemble quickly at the beginning of the stream because concept drift detection triggers the reset of the window size and speeds up the introduction of new models. \axgbpush and \axgbrepl fill the ensemble at a slower rate and finish filling the ensemble before the 200K mark. This is in line with our expectations given the introduction of new models trained on increasing window sizes as defined in \refequation{eq:window_size}. Finally, \axgbarepl is the slowest to fill the ensemble at around the 700K mark. This is expected given that upon drift detection, \axgbarepl starts replacing the oldest models of the ensemble.

It is important to mention that additional memory resources are used by the different \axgb variants given their batch-incremental nature: mini-batches are accumulated in memory before they are used to fit a tree. In this sense, other things being equal, instance-incremental methods are more memory efficient. However, our results show that all versions of \axgb keep the size of the model under control, a critical feature when facing theoretically infinite data streams.

\subsection{Training Time }\label{sec:results_time}
\begin{figure}[!t]
    \centering
    \begin{subfigure}[b]{0.4\textwidth}
        \includegraphics[width=\textwidth]{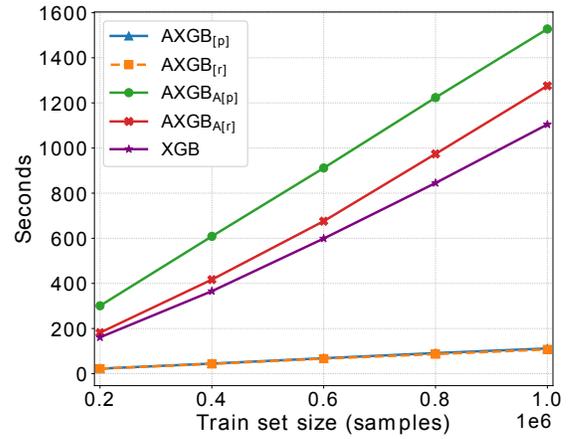}
        \label{fig:time}
    \end{subfigure}
    \\
    \begin{subfigure}[b]{0.4\textwidth}
        \includegraphics[width=\textwidth]{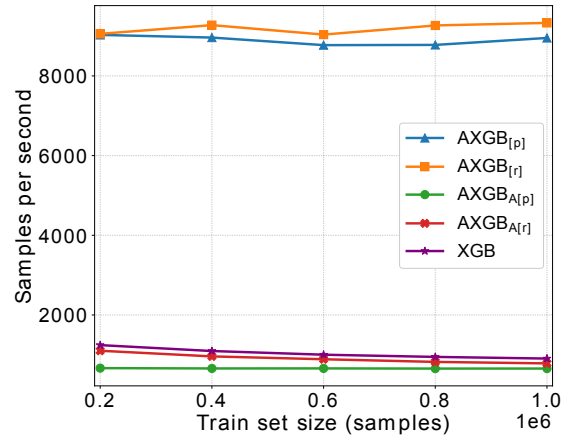}\label{fig:throughput}
    \end{subfigure}
    \caption{Training time (top) and throughput (bottom) test results.}\label{fig:time_test}
\end{figure}
Finally, we measure training time for the different versions of \axgb. We use as reference the time required to train an \xgb model on the \textit{Madelon} dataset used in the model complexity test. Models are trained using the following configuration: ensemble size = 30, max window size = 10K, min window size = 1, learning rate = 0.05, max depth =~6.
We used the following dataset sizes: 200K, 400K, 600K, 800K and 1M. Results correspond to the average time after running the experiments 10 times for each dataset size and for each classifier. Measurements are shown in \reffig{fig:time_test} in terms of time (seconds) and in \reffig{fig:time_test} in terms of throughput (samples per second). These tests show that the fastest learners are \axgbpush and \axgbrepl, both showing small change in training time as the number of instances increases. This is an important feature given that 
training time plays a key role in stream learning applications. On the other hand, \axgbapush and \axgbarepl have similar behaviour in terms of training time compared to \xgb while being slightly slower. This can be attributed to the overhead from the drift-detection process, which implies getting predictions for each instance and keeping the drift detector statistics. Additionally, we see that \axgbapush is the slowest classifier, which might be related to the overhead incurred by predicting using more ensemble members, given that the ensemble is quickly filled as previously discussed. 

\section{Conclusions}\label{sec:conclusions}
In this paper, we propose an adaptation of the eXtreme Gradient Boosting algorithm~(\xgb) to evolving data streams. The core idea of \textsc{Adaptive XGBoost}~(\axgb) is the incremental creation/update of the ensemble, i.e., weak learners are trained on mini-batches of data and then added to the ensemble. We study variations of the proposed method by considering two main factors: concept drift awareness and the strategy to update the ensemble. We test \axgb against instance-incremental and batch-incremental methods on synthetic and real-world data. Additionally, we consider a simple batch-incremental approach~(\bxgb) consisting of ensemble members that are full \xgb models trained on consecutive mini-batches. From our tests, we conclude that \axgbrepl (the version that performs model replacement in the ensemble but does not include explicit concept drift awareness) represents the best compromise in terms of performance, training time and memory usage. 

Another noteworthy finding from our experiments is the good predictive performance of \bxgb after parameter tuning. If resource consumption is a secondary consideration, this approach may be a worthwhile candidate for application in practical data stream mining scenarios, particularly considering that our parameter tuning experiments did not investigate optimizing the size of the boosted ensemble for each mini-batch in \bxgb. (The size of each sub-ensemble was fixed at 30 members.) Overall, despite the limitations of mini-batch-based data stream mining, and its drawbacks compared to instance-incremental methods, it appears that \xgb-based techniques are promising candidates for data stream applications. In a similar way, we believe \axgb is an interesting alternative to \xgb for some applications given its efficient management of resources and adaptability.


%
%
\bibliographystyle{IEEEtran}
\bibliography{thebibliography.bib}
\end{document}